\documentclass[11pt]{article}

\usepackage[final]{acl}

\usepackage{times}
\usepackage{latexsym}

\usepackage[T1]{fontenc}

\usepackage[utf8]{inputenc}

\usepackage{microtype}

\usepackage{inconsolata}

\usepackage{graphicx}
\usepackage{booktabs}
\usepackage{tabularx}
\usepackage{array} 
\usepackage[table]{xcolor}
\usepackage{multirow}

\usepackage{float}
\usepackage{tcolorbox}
\usepackage{xcolor}
\usepackage{stfloats}
\usepackage{hyperref}

\usepackage{enumitem}
\usepackage[normalem]{ulem}
\usepackage{amsmath}

\title{MedRCube: A Multidimensional Framework for Fine-Grained and In-Depth Evaluation of MLLMs in Medical Imaging}

\author{
    \textbf{Zhijie Bao\textsuperscript{\rm 1, \rm 2}\footnotemark[2]},
    \textbf{Fangke Chen\textsuperscript{\rm 2, \rm 3}}\footnotemark[2],
    \textbf{Licheng Bao\textsuperscript{\rm 1}},
    \textbf{Chenhui Zhang\textsuperscript{\rm 1}},
    \\
    \textbf{Wei Chen\textsuperscript{\rm 4}},
    \textbf{Jiajie Peng\textsuperscript{\rm 5}\footnotemark[1]},
    \textbf{Zhongyu Wei\textsuperscript{\rm 1, \rm 2}\footnotemark[1]}
    \\
    \normalsize{\textsuperscript{1}School of Data Science, Fudan University, China,}
    \\
    \normalsize{\textsuperscript{2}Shanghai Innovation Institute, China,}
    \\
    \normalsize{\textsuperscript{3}School of Integrated Circuits, Zhejiang University, China,}
    \\
    \normalsize{\textsuperscript{4}School of Software Engineering, Huazhong University of Science and Technology, China,}
    \\
    \normalsize{\textsuperscript{5}School of Computer Science, Northwestern Polytechnical University, China,}
    \\
    \normalsize{\texttt{zjbao24@m.fudan.edu.cn, fkchen@zju.edu.cn}}\\
    \normalsize{\texttt{jiajiepeng@nwpu.edu.cn, zywei@fudan.edu.cn}}
    \\
}

\begin{document}
\maketitle
\begin{abstract}


The potential of Multimodal Large Language Models (MLLMs) in domain of medical imaging raise the demands of systematic and rigorous evaluation frameworks that are aligned with the real-world medical imaging practice. Existing practices that report single or coarse-grained metrics are lack the granularity required for specialized clinical support and fail to assess the reliability of reasoning mechanisms. To address this, we propose a paradigm shift toward multidimensional, fine-grained and in-depth evaluation. Based on a two-stage systematic construction pipeline designed for this paradigm, we instantiate it with MedRCube. We benchmark 33 MLLMs, \textit{Lingshu-32B} achieve top-tier performance. Crucially, MedRCube exposes a series of pronounced insights inaccessible under prior evaluation settings. Furthermore, we introduce a credibility evaluation subset to quantify reasoning credibility, uncover a highly significant positive association between shortcut behavior and diagnostic task performance, raising concerns for clinically trustworthy deployment. The resources of this work can be found at \url{https://github.com/F1mc/MedRCube}.
\end{abstract}



\section{Introduction}

\footnotetext[2]{The two authors contribute equal to this work.}
\footnotetext[1]{Corresponding authors.}

Multimodal large language models (MLLMs) have recently demonstrated strong performance in medical imaging tasks, indicating their potential for a wide range of radiology-related applications~\cite{fang2025large,chen2024towards,bluethgen2025best,yang2024advancing}. At the same time, emerging studies and reports suggest that clinicians are beginning to use MLLMs as decision-support tools and to incorporate them into radiology workflows~\cite{hou2025one, wada2025retrieval}. Given the safety-critical nature of clinical decision-making, this development highlights the need for systematic and rigorous evaluation frameworks that are aligned with the requirements and constraints of real-world medical imaging practice.

To evaluate the capabilities of multimodal large language models (MLLMs) in medical imaging tasks, several medical visual question answering (VQA) benchmarks have been introduced, such as VQA-RAD and SLAKE~\cite{vqarad,slake}. These benchmarks provide a standardized setting for model evaluation and comparison. In practice, existing evaluation protocols commonly report model performance using a single aggregate score, or adopt coarse-grained metrics defined along dimensions such as imaging modality or anatomical region~\cite{vqarad,omnimedvqa,gmai}. 

\begin{figure}[t]
  \centering
  \includegraphics[width=0.8\linewidth]{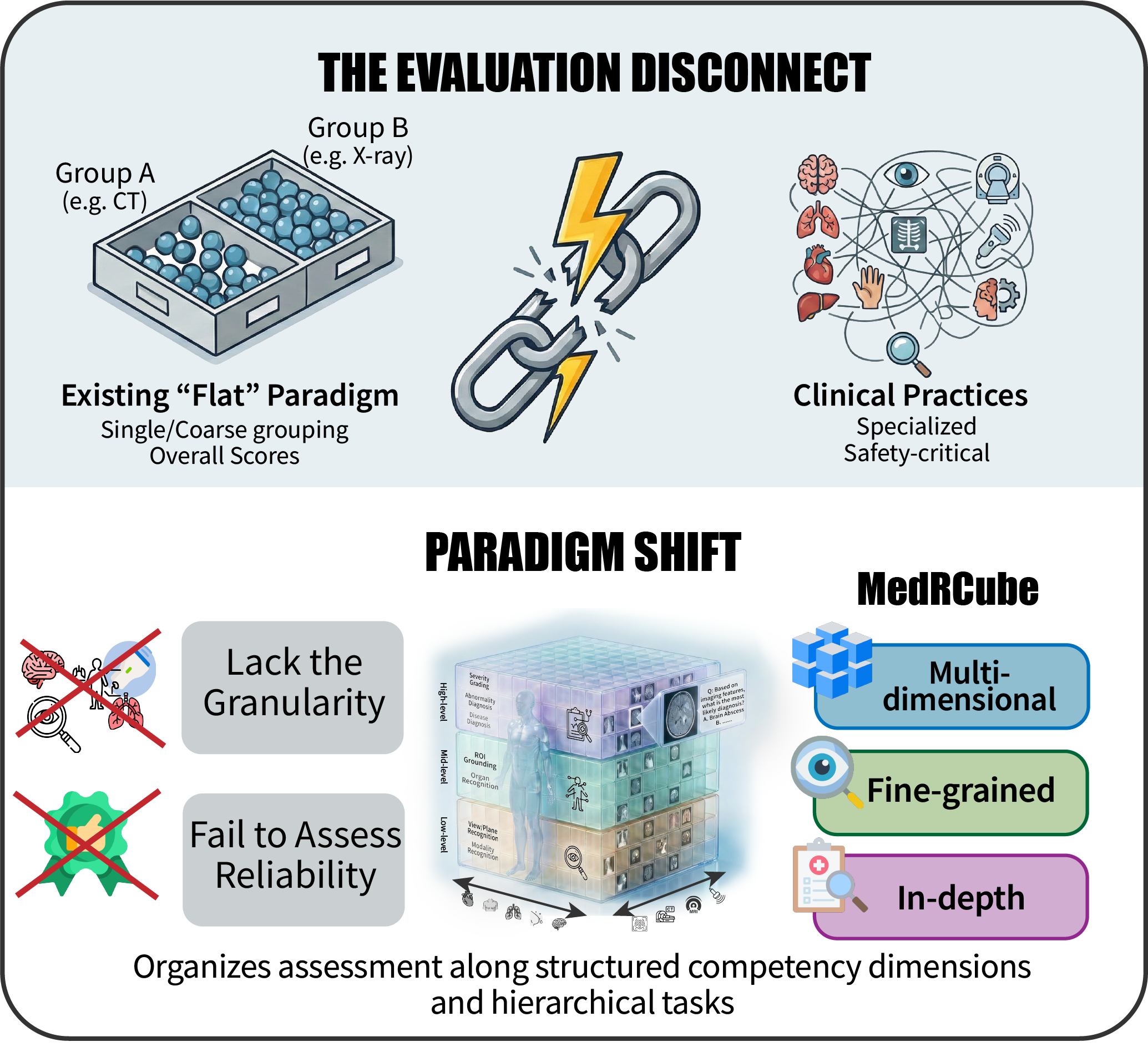}
  \vspace{-0.2cm}
  \caption{\textbf{Paradigm shift} required for evaluating multimodal large language models in medical imaging.}
  \vspace{-0.3cm}
  \label{fig:intro}
\end{figure}

While valuable, current evaluations (i) \emph{lack the granularity required for specialized clinical support}. Since healthcare is strictly compartmentalized, aggregate scores (e.g., general magnetic resonance imaging performance) offer little assurance to specialists like cardiologists, who need validation on specific pathologies rather than broad metrics. More critically, existing protocols (ii) \emph{fail to assess the reliability of reasoning mechanisms}. As clinical diagnosis follows a strict logical hierarchy, current metrics cannot distinguish valid reasoning from "hallucinated correctness", where a model outputs the right diagnosis without correctly perceiving the target organ. Such evidence-free decisions are inadmissible in rigorous clinical practice. These two limitations reveal the gap between the current evaluation framework and the need in clinical scenarios and require a paradigm shift. We therefore argue for a multidimensional evaluation framework that reflects the inherent complexity of medical imaging by transforming assessment from a linear list of metrics into a dense, fine-grained competency space. This structure enables granular analysis, allowing researchers to disaggregate performance data to identify specific capability profiles and verify the validity of reasoning mechanisms---a prerequisite for trustworthy clinical deployment.

However, implementing this framework presents significant challenges. Such fine-grained analysis requires comprehensive coverage of the structured competency space, yet available resources remain loosely organized and unevenly distributed~\cite{projectimagingx2025}. Furthermore, the requirement for clinical reliability demands a standard of item rigor and quality that significantly exceeds previous benchmarks~\cite{vydareny1986guidelines,gunabushanam2019automated}. To address these challenges, we developed a two-stage framework to transform fragmented data into a structured competency space, enabling the large-scale generation of rigorous items. Throughout the framework, we ground complementary clinical knowledge resources (e.g., HPO, RadLex, ICD-11) and clinician verification to ensure clinical rigor and consistency~\cite{kohler2021human, rsnaradlex, whoicd11}.


Building on this framework, we present \textbf{MedRCube}, a multidimensional medical imaging benchmark designed for the fine-grained and in-depth evaluation of MLLMs. It organizes evaluation across three dimensions: \textbf{Anatomical Region} (5 types), \textbf{Imaging Modality} (4 types) and \textbf{Task Type} (8 tasks). Notably, we structure the task dimension into a cognitive hierarchy covering perception, semantics, and cognition to align with clinical reasoning processes. In total, MedRCube comprises 7,626 samples curated from 36 datasets.

Beyond coarse ranking, MedRCube enables analyses addressing the following three research questions:
\begin{itemize}[itemsep=2pt, topsep=2pt, parsep=1pt]
    \item \textbf{RQ0 (Overall capability profiling).} How do MLLMs perform across the competency axes of anatomy, modality, and task hierarchy?
    \item \textbf{RQ1 (Skill factors and relationships).} 
    What underlying capability factors or relationship patterns emerge across the structured competency space from a cross-model perspective?
    \item \textbf{RQ2 (Credibility and shortcut behavior).} Do models exhibit grounded reasoning or rely on shortcuts? Specifically, do they produce correct high-level answers despite failing prerequisite tasks, and how can this behavior be quantified?
\end{itemize}

Using MedRCube, we benchmark \textbf{33 MLLMs}.
Beyond overall ranking, MedRCube yields insight largely inaccessible to flat metrics, including a "foundational blindness" that reveals models lack robust mastery of basic but rare perceptual primitive, a pronounced "brain island" effect revealing the isolation of brain-related capabilities, and a highly significant positive association between shortcut behavior and high-level performance. This indicates that stronger models show more opportunistic reliance on shortcuts in high-level tasks.

\begin{figure*}
    \centering
    \includegraphics[width=\textwidth]{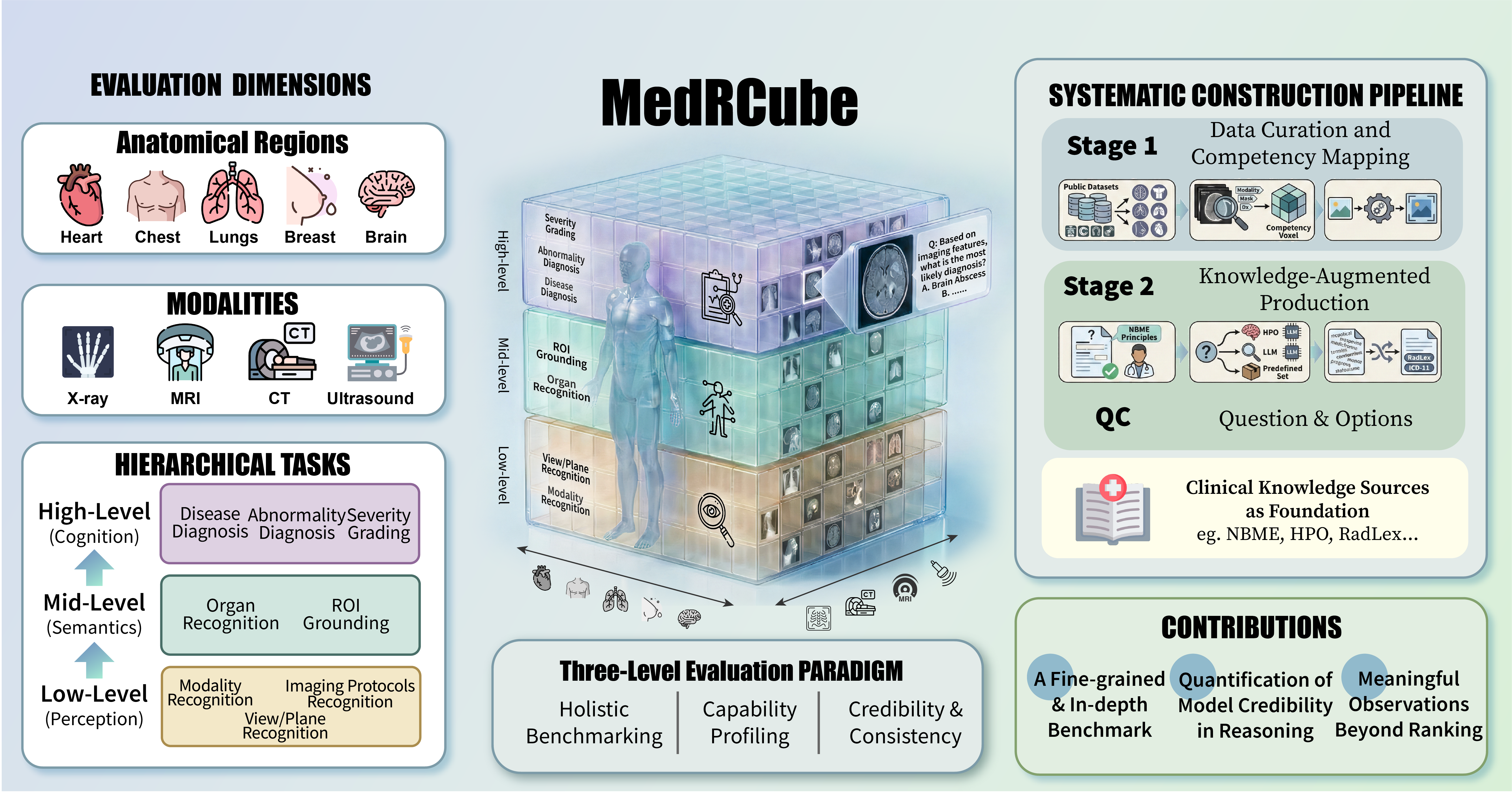}
    \vspace{-0.2cm}
    \caption{Overview of MedRCube. The framework evaluates models along Anatomical, Modality, and Task Hierarchy axes. The pipeline (right) aligns datasets via Stage I and generates rigorous items via Stage II with expert verification.}
    \vspace{-0.3cm}
    \label{fig:main}
\end{figure*}

\section{The MedRCube Framework: Concept and Design}
\subsection{Overview}
To address the limitations of flat metrics, we introduce a multidimensional evaluation framework designed for fine-grained capability profiling and in-depth analysis. This framework organizes evaluation units into a dense \textbf{competency space} defined by anatomical regions, imaging modalities, and a cognitive task hierarchy. This structured design moves beyond aggregate scoring, enabling a \textbf{three-level in-depth evaluation} that assesses both granular capability and reasoning credibility.

\subsection{Competency Space} \label{subsec:design} The core of MedRCube is its multi-dimensional taxonomy. We conceptually model the benchmark as a structured space, named as the Competency Space along three orthogonal axes:

\textbf{Anatomical \& Modality Axes.} To ensure broad clinical coverage, we encompass five anatomical regions, including the Heart, Chest, Breast, Lung, and Brain, along with four imaging modalities, namely X-ray, Computed Tomography (CT), Magnetic Resonance Imaging (MRI), and Ultrasound.

\textbf{Task Hierarchy Axis.} Crucially, we structure tasks not as a flat list, but as a \textbf{cognitive hierarchy} mirroring the radiological logic. This allows us to trace model failures from basic perception to complex reasoning: 
\begin{itemize}[itemsep=2pt, topsep=3pt, parsep=1pt]
    \item Low-level (Perception): Modality recognition, view/plane recognition and imaging protocols recognition.
    \item Mid-level (Semantics): Organ recognition and Region-of-Interest (ROI) grounding.
    \item High-level (Cognition): Abnormality diagnosis, disease diagnosis, and severity grading.
\end{itemize}
Detailed definitions and illustrative examples for each task are provided in Appendix~\ref{app:samplesexamples}.

\textbf{Competency Voxel.}
Within this competency space, samples sharing the same intersection of coordinates (Anatomy, Modality, Task) constitute a fundamental evaluation unit, which we term a Competency Voxel. This granular aggregation allows for precise localization of model capabilities and deficits.

\subsection{In-depth Evaluation Paradigm}
\label{subsec:diagnostic_paradigm}

Based on the proposed multidimensional and fine-grained framework, MedRCube enables a comprehensive diagnostic paradigm that progresses from holistic ranking to fine-grained capability mining and cognitive consistency verification. This paradigm is structured into three levels:

\textbf{Holistic Benchmarking.}
First and foremost, MedRCube serves as a rigorous standard for measuring the overall competence of MLLMs. MedRCube leverages a balanced distribution across anatomical regions and imaging modalities to ensure the validity of assessment.

\textbf{Capability Profiling.}
Going beyond global rankings, the multidimensional design allows for multi-dimensional performance analysis. We can slice model capabilities along specific axes or zoom into specific competency voxels to mine granular failure modes, which are typically obscured in flat evaluation metrics.

\textbf{Credibility and Consistency.}
The proposed cognitive hierarchy provides a novel perspective to define and quantify the credibility of model reasoning. By verifying whether high-level diagnoses are supported by prerequisite perceptual success on the \emph{same} image, we expose "correct-by-accident" shortcuts---where models reach right answers despite failing basic recognition (e.g. give the correct diagnosis but cannot even tell it is an MRI image)---thereby distinguishing grounded reasoning from spurious guessing.

\section{Systematic Construction Pipeline}
\label{sec:pipeline}
Unlike conventional VQA datasets prioritizing maximizing scale or task diversity in isolation, MedRCube must densely populate a structured competency space while preserving clinical determinism, terminological consistency, and hierarchical validity across tasks.
To this end, we implement a systematic pipeline consisting of two stages: metadata-driven competency mapping (Stage I) to align fragmented sources, and knowledge-augmented item production (Stage II) to generate rigorous items, underpinned by a dedicated quality assurance procedure. Design and iteration of pipeline is completed under the deep involvement and strict supervision of radiologists. An overview of the pipeline is illustrated in Fig.~\ref{fig:main}. What’s more, for details regarding this chapter, please refer to Appendix~\ref{app:designdetails}.

\subsection{Stage I: Data Curation and Competency Mapping}
\paragraph{Data Collection.}



We curate 35 datasets (detailed in Table~\ref{tab:medical_datasets}) spanning diverse acquisition protocols. To mitigate heterogeneity, we apply a unified processing pipeline for image and spatial annotations (masks/boxes) and disregard the original dataset-specific task labels. Instead, we treat all samples as raw clinical evidence, re-interpreting them strictly under MedRCube's standardized taxonomy to ensure consistent evaluation.

\paragraph{Metadata-Driven Task Derivation.}
Most datasets are not explicitly annotated for the full spectrum of perceptual, semantic, and cognitive tasks required by the competency cube. To address this, we adopt a metadata-driven task derivation strategy.

For each image, we therefore derive task support from available metadata (e.g., modality/acquisition tags for perceptual tasks, segmentation masks for semantic tasks, and diagnostic annotations for cognitive tasks) and assign each sample to one or more Competency Voxels, ensuring that task assignments are grounded in available evidence rather than heuristic dataset labels.



\subsection{Stage II: Knowledge-Augmented Production}
While Stage I aligns images to the dense competency space, Stage II focuses on producing clinically rigorous and diagnostically meaningful items at scale. From question formulation to option construction, the entire production process is explicitly guided by National Board of Medical Examiner (NBME)~\cite{nbme2021item} item-writing principles, which we adopt as a unified design specification throughout the benchmark.

\paragraph{Expert-verified Question Design.}
For each task type, we design a pool of question templates aligned with item-writing principles. Templates are crafted to ensure that:
(i) the question can be answered solely based on the image and prompt,
(ii) the question targets a single, well-defined competency, and
(iii) the correct answer is clinically deterministic. All templates undergo clinician review to verify clinical relevance, clarity, and task alignment.
All question designs are verified by experts to ensure they are well-formed and answerable.

\paragraph{Knowledge-Augmented Distractor Generation.}
We stratify questions into three distinct answer-space types: disease-oriented, abnormality-oriented, and closed-set. For each type, we employ task-aware, knowledge-augmented strategies to ensure distractors are clinically plausible and strictly difficulty-controlled. Disease-oriented questions leverage ontology-guided retrieval (e.g., HPO) followed by controlled LLM generation, while other types employ constrained LLM generation or predefined and verified candidate pools.

\paragraph{Terminology Standardization.}
To ensure terminological consistency across datasets and tasks, all medical terms appearing in questions and options are normalized to authoritative clinical vocabularies. Radiological findings and anatomical structures are standardized using RadLex, while disease entities are mapped to ICD-11 names.




\subsection{Quality Assurance}
To ensure clinical validity, all items undergo a rigorous NBME-guided review. Question Review ensures items are unambiguous and clinically deterministic, meaning the correct answer is derivable solely from the image without relying on options. Option Review validates that distractors are conceptually homogeneous, mutually exclusive, and clinically plausible within the context. We use the LLM-based method to take this review and review its reliability, the consistency of annotations from LLM and experts is above 90\% on all rules we used, \ref{subsec:qualityreview} reports details.

\begin{table*}[t!]
\centering
\caption{Performance comparison on MedRCube. The best scores are highlighted in bold, and the second-best are underlined. All scores are reported in percentage (\%). Abbreviations for subtasks and corresponding full terminology are detailed in the Table \ref{tab:abbre}.}
\label{tab:results}
\definecolor{graybg}{gray}{0.9}
\definecolor{bluebg}{rgb}{0.88, 0.95, 1.0}
\definecolor{cyanbg}{rgb}{0.88, 1.0, 1.0}
\resizebox{\textwidth}{!}{%
\begin{tabular}{l|c|cccccccc|cccc|ccccc}
\toprule
\multirow{2}{*}{\textbf{Model}} & \multirow{2}{*}{\textbf{Overall}} & \multicolumn{8}{c|}{\textbf{Task Type}} & \multicolumn{4}{c|}{\textbf{Modality Type}} & \multicolumn{5}{c}{\textbf{Part Type}} \\
\cmidrule(lr){3-10} \cmidrule(lr){11-14} \cmidrule(lr){15-19}
 & & MR & VR & IPR & OR & RG & DD & AD & SG & CT & X-Ray & MRI & US & Brain & Chest & Heart & Lungs & Breast \\
\midrule
Random & 25.64 & 24.22 & 23.67 & 24.57 & 24.37 & 23.00 & 28.63 & 21.40 & 28.52 & 24.65 & 27.48 & 24.61 & 22.96 & 26.41 & 26.49 & 22.62 & 24.55 & 29.13 \\
\midrule
\rowcolor{graybg} 
\multicolumn{19}{c}{\textit{\textbf{Proprietary MLLMs}}} \\
\midrule
Qwen3-VL-Plus & 51.14 & 82.66 & 50.00 & 23.14 & 70.01 & 42.06 & 47.03 & 30.74 & \textbf{44.72} & 60.54 & 59.81 & 37.63 & 40.45 & 38.81 & 62.70 & 45.71 & 54.79 & 46.32 \\
GPT-5.1 & 53.36 & 85.66 & 61.00 & 33.71 & 74.70 & 35.20 & 52.84 & 21.79 & 37.50 & 69.61 & 58.86 & 43.29 & 34.62 & 41.33 & 62.01 & 53.33 & 62.18 & 38.42 \\
Claude Opus 4.5 & 53.38 & 85.37 & \underline{64.33} & 33.29 & 73.76 & 54.30 & 44.05 & 20.23 & \underline{44.54} & 61.86 & 61.34 & 43.42 & 35.34 & 38.49 & 63.79 & 59.05 & 54.79 & 42.16 \\
Gemini-3-Pro & 59.35 & 83.53 & \textbf{71.33} & \underline{37.86} & \underline{84.25} & 45.30 & 57.91 & \underline{42.02} & 35.39 & 71.94 & 62.21 & 64.21 & 42.87 & 51.59 & 64.70 & \textbf{88.81} & 66.26 & 37.45 \\
\midrule
\rowcolor{bluebg}
\multicolumn{19}{c}{\textit{\textbf{Open-Source MLLMs}}} \\
\midrule
InternVL3.5-2B & 51.20 & 88.86 & 49.67 & 26.71 & 61.95 & 38.40 & 50.51 & 32.10 & 36.44 & 68.53 & 54.00 & 61.45 & 31.30 & 45.81 & 57.04 & 49.76 & 69.95 & 32.32 \\
LLaVa-v1.5-7B & 39.22 & 68.99 & 6.00 & 7.43 & 69.79 & 31.80 & 36.56 & 19.26 & 28.87 & 47.52 & 43.28 & 60.13 & 31.66 & 39.38 & 44.39 & 45.95 & 55.54 & 32.87 \\
Phi-3.5-Vision & 36.14 & 74.32 & 9.00 & 24.43 & 49.77 & 27.90 & 31.00 & 13.23 & 27.29 & 42.25 & 43.25 & 25.53 & 28.88 & 26.36 & 43.27 & 18.57 & 44.93 & 41.19 \\
Janus-Pro-7B & 40.63 & 79.84 & 22.67 & 32.29 & 63.07 & 39.40 & 26.99 & 18.87 & 27.64 & 44.26 & 47.96 & 41.45 & 23.23 & 24.43 & 49.70 & 56.19 & 48.53 & 24.69 \\
Qwen2.5-VL-7B & 46.34 & 90.41 & 1.67 & 9.86 & 66.17 & 42.00 & 45.11 & 14.79 & 39.08 & 58.91 & 53.22 & 49.61 & 43.14 & 39.38 & 54.86 & 52.86 & 56.97 & \textbf{55.06} \\
Qwen3-VL-8B & 55.12 & \underline{91.09} & 43.33 & 29.86 & 73.95 & 40.40 & 55.50 & 24.51 & 43.66 & 69.15 & 59.09 & 56.97 & 44.30 & 49.15 & 61.86 & 61.67 & 65.31 & 46.05 \\
InternVL3-8B & 49.43 & 85.17 & 53.33 & 32.00 & 70.48 & 52.60 & 35.50 & 28.21 & 38.03 & 53.02 & 55.82 & 46.18 & 37.58 & 37.84 & 58.29 & 55.00 & 49.29 & 42.16 \\
InternVL3.5-8B & 55.87 & 85.66 & 55.00 & 28.00 & 74.60 & 52.10 & 51.86 & 39.30 & 40.32 & 58.76 & 60.94 & 61.97 & 50.40 & 54.60 & 63.89 & 61.43 & 56.97 & 45.08 \\
MiniCPM-o 2.6 & 41.12 & 78.10 & 32.67 & 31.14 & 51.64 & 38.80 & 29.16 & 37.74 & 29.58 & 46.90 & 43.17 & 35.66 & 40.36 & 40.03 & 45.36 & 42.14 & 43.89 & 32.73 \\
\multicolumn{19}{l}{\textit{------ 10B Level Boundary ------}} \\
InternVL3-14B & 49.95 & 79.26 & 40.67 & 29.14 & 66.64 & \underline{54.50} & 42.86 & 31.91 & 34.68 & 56.51 & 59.32 & 43.68 & 33.09 & 44.99 & 62.29 & 37.38 & 50.90 & 33.56 \\
Llama-3.2-11B-Vision & 40.56 & 66.86 & 49.33 & 32.71 & 57.92 & 39.20 & 31.21 & 11.09 & 34.51 & 39.07 & 46.40 & 38.29 & 28.25 & 29.21 & 48.02 & 41.67 & 37.06 & 35.23 \\
Qwen2.5-VL-32B & 53.06 & 85.85 & 26.33 & 32.86 & 74.88 & 53.80 & 47.20 & 26.46 & 39.44 & 70.00 & 60.01 & 45.13 & 37.13 & 44.51 & 62.73 & 49.29 & 66.45 & 38.00 \\
Qwen3-VL-32B & 54.43 & 88.47 & 56.00 & 33.71 & 78.44 & 43.30 & 48.38 & 29.96 & 39.96 & 60.93 & 60.73 & 49.74 & 43.14 & 42.72 & 63.67 & 64.52 & 54.98 & 46.19 \\
InternVL3.5-14B & 57.28 & 89.05 & 53.33 & 31.43 & 72.82 & 51.80 & 57.83 & 25.10 & 39.61 & 71.16 & \underline{63.77} & 56.71 & 38.21 & 44.51 & 66.57 & 55.00 & 71.84 & 43.69 \\
InternVL3.5-38B & 58.14 & 87.02 & 58.00 & 31.43 & 74.88 & \textbf{55.90} & 59.39 & 21.21 & 39.26 & \underline{75.58} & \textbf{64.20} & 59.87 & 34.80 & 42.72 & \underline{67.60} & 69.05 & \textbf{74.88} & 37.59 \\
\midrule
\rowcolor{cyanbg}
\multicolumn{19}{c}{\textit{\textbf{Medical-specific MLLMs}}} \\
\midrule
MedVLM-R1 & 43.61 & 86.92 & 25.67 & 27.86 & 67.67 & 41.50 & 29.53 & 20.62 & 33.80 & 48.14 & 50.36 & 48.68 & 28.70 & 26.53 & 52.39 & 62.86 & 50.71 & 34.95 \\
HealthGPT-M3 & 42.02 & 49.52 & 42.33 & 34.29 & 63.64 & 33.90 & 38.77 & 36.38 & 30.46 & 57.91 & 42.04 & 27.11 & 38.48 & 40.85 & 44.08 & 27.38 & 60.00 & 24.41 \\
LLaVA-Med-7B & 24.51 & 22.00 & 31.33 & 23.71 & 21.65 & 31.20 & 23.56 & 20.04 & 28.17 & 19.61 & 31.72 & 18.95 & 10.22 & 20.02 & 31.37 & 17.14 & 15.73 & 16.78 \\
HuatuoGPT-Vision-7B & 47.11 & 89.73 & 40.33 & 12.86 & 74.79 & 42.00 & 39.02 & 13.81 & 37.50 & 62.48 & 54.29 & 48.29 & 29.60 & 32.79 & 56.89 & 60.48 & 60.57 & 36.75 \\
Lingshu-7B & \underline{59.86} & \textbf{91.38} & 53.33 & 30.57 & 82.29 & 42.40 & \underline{65.69} & 26.26 & 36.09 & 74.26 & 58.94 & \textbf{72.76} & \textbf{57.40} & \underline{63.79} & 62.26 & 68.33 & 70.62 & \underline{52.98} \\
Hulu-Med-4B & 48.95 & 79.55 & 50.67 & 31.57 & 63.73 & 30.80 & 44.62 & \textbf{43.77} & 41.37 & 60.16 & 51.40 & 42.24 & 43.41 & 38.24 & 53.58 & 55.71 & 57.44 & 46.46 \\
Hulu-Med-7B & 42.14 & 61.24 & 54.33 & 34.00 & 55.30 & 30.10 & 39.06 & 26.65 & 34.86 & 48.76 & 48.43 & 34.61 & 21.97 & 27.91 & 50.52 & 43.10 & 46.92 & 24.55 \\
\multicolumn{19}{l}{\textit{------ 10B Level Boundary ------}} \\
HealthGPT-L14 & 44.43 & 76.07 & 45.33 & 25.00 & 60.07 & 37.60 & 38.73 & 20.43 & 34.68 & 56.74 & 53.34 & 32.11 & 20.54 & 27.91 & 56.73 & 42.14 & 50.71 & 24.97 \\
HealthGPT-XL32 & 47.13 & 72.67 & 43.67 & 32.71 & 68.32 & 38.60 & 39.75 & 35.60 & 37.68 & 54.88 & 55.71 & 28.42 & 34.26 & 37.99 & 58.79 & 36.90 & 49.19 & 29.26 \\
HuatuoGPT-Vision-34B & 54.02 & 89.15 & 56.33 & 31.00 & 78.44 & 33.40 & 53.66 & 21.40 & 38.91 & 63.10 & 56.14 & 62.63 & 44.93 & 47.19 & 58.70 & 60.00 & 63.32 & 49.24 \\
MedDr-40B & 28.44 & 29.17 & 27.67 & 27.29 & 28.11 & 25.80 & 30.55 & 27.43 & 26.06 & 27.13 & 30.25 & 24.74 & 27.80 & 25.79 & 30.18 & 27.14 & 26.45 & 30.37 \\
Lingshu-32B & \textbf{62.55} & 90.31 & 53.67 & 37.00 & \textbf{84.72} & 49.90 & \textbf{65.77} & 40.47 & 35.03 & \textbf{77.83} & 63.68 & \underline{70.00} & \underline{54.71} & \textbf{65.42} & \textbf{67.73} & \underline{69.76} & \underline{73.27} & 43.27 \\
Hulu-Med-14B & 50.28 & 72.29 & 59.00 & 36.57 & 68.88 & 39.20 & 47.53 & 28.40 & 38.73 & 62.79 & 56.43 & 46.71 & 25.38 & 37.02 & 60.07 & 54.76 & 58.29 & 24.69 \\
Hulu-Med-32B & 55.27 & 82.56 & 62.00 & \textbf{40.71} & 69.35 & 45.90 & 52.88 & 31.71 & 41.73 & 67.36 & 61.20 & 57.76 & 28.52 & 43.37 & 64.67 & 59.29 & 64.36 & 29.54 \\
\bottomrule
\end{tabular}%
}
\end{table*}

\section{Experiments}

\subsection{Settings}
We evaluate \textbf{33} models on MedRCube, comprising 4 proprietary models, 14 Medical MLLMs, and 15 General-purpose MLLMs. For detailed information on the evaluation process and the models being evaluated, please refer to the Appendix~\ref{app:evaluationdetails}. We additionally report a \textbf{text-only} baseline (questions/options without images) to quantify potential prior leakage; most models drop to near random-guessing performance (Appendix~\ref{app:textonly}).
\subsection{Results}
\label{subsec:mainresults}
Table \ref{tab:results} reports the evaluation results of 33 models on MedRCube.

\textbf{Overall performance and ranking.}
\textit{Lingshu-32B} achieves the best performance (62.55), followed by \textit{Lingshu-7B} (59.86), \textit{Gemini-3-Pro} (59.35), and \textit{InternVL3.5-38B} (58.14). Notably, the top-performing models span both proprietary and open-source paradigms, as well as general-purpose and medically specialized training regimes. At the same time, a large fraction of models—including several medically fine-tuned systems—cluster in a relatively narrow performance band (approximately 45 - 55), indicating that flat aggregate metrics compress substantial underlying heterogeneity.

\textbf{Task hierarchy.}
Performance across task types follows the intended cognitive gradient, inversely correlating with cognitive complexity. Crucially, the advantage of specialized MLLMs peaks in high-level reasoning tasks, while gaps remain marginal in saturated perceptual tasks.

\textbf{Modalities.}
CT and X-ray tasks are generally the strongest, whereas Ultrasound remains consistently challenging even for top-performing models. MRI exhibits high variance across models, reflecting sensitivity to training exposure or model design.

\textbf{Anatomical regions.}
Performance also varies by anatomical region. The Chest, lung, and heart tasks are generally better handled than the remains. We observe that medical specialization is most likely to translate into gains on brain and breast imaging.

\subsection{Key Findings and Analysis}
\textbf{Correlation with Data Distribution.}
Cross-referencing model performance with a medical imaging datasets survey~\citep{projectimagingx2025}, we observe \uline{a strong alignment between open-source data availability and model proficiency}. Well-optimized models (e.g., \textit{Lingshu}, \textit{InternVL}) consistently excel in high-resource domains such as CT, X-ray, and Brain imaging (ranked among the top five in data availability), while performance degrades in lower-resource modalities like Ultrasound. 

\uline{A notable exception is Breast imaging}: despite substantial dataset representation ($\sim$14.9\%), performance remains disproportionately low across models, likely due to inherent task difficulty or data quality.

\textbf{Weakened Scaling Effect.}
Analyzing maximum scores within both general-purpose and medical MLLMs, we find that large models ($>$10B) fail to establish a decisive advantage over smaller counterparts ($<$10B), with performance gaps narrowing to under 2.5\%. This indicates that current data scales may not fully activate scaling laws in medical imaging VQA.

In contrast, domain adaptation proves critical: the lightweight medical model \textit{Lingshu-7B} ($63.25\%$) outperforms the much larger, top-ranked generalist \textit{InternVL3.5-38B} ($60.97\%$), suggesting that \uline{targeted medical training outweighs raw parameter expansion}.

\textbf{Blind Spots in Basic Perceptual Skills.}
Despite being fundamental competencies for human radiologists, \uline{low-level perceptual tasks that are rarely emphasized in existing datasets exhibit surprisingly weak performance}. In particular, imaging protocol recognition (e.g., distinguishing T1 vs.\ T2 MRI, rather than MRI vs.\ other modalities) remains extremely challenging, with most models scoring only in the 20 - 40\% range. 

Similarly, while the view recognition task is trivial for radiologists and specialized discriminative models (near-perfect accuracy)~\cite{rajkomar2017high}, current benchmarks reveal a significant gap: even the strongest open-source models reach only $62\%$. These results reveal a critical blind spot — ~\uline{models lack robust mastery of basic but rare perceptual primitives}, which is typically assumed as prerequisites for higher-level clinical reasoning.

\section{Correlation Analysis}
\begin{figure}[t!]
    \centering
    \scalebox{1}[0.9]{\includegraphics[width=0.95\linewidth]{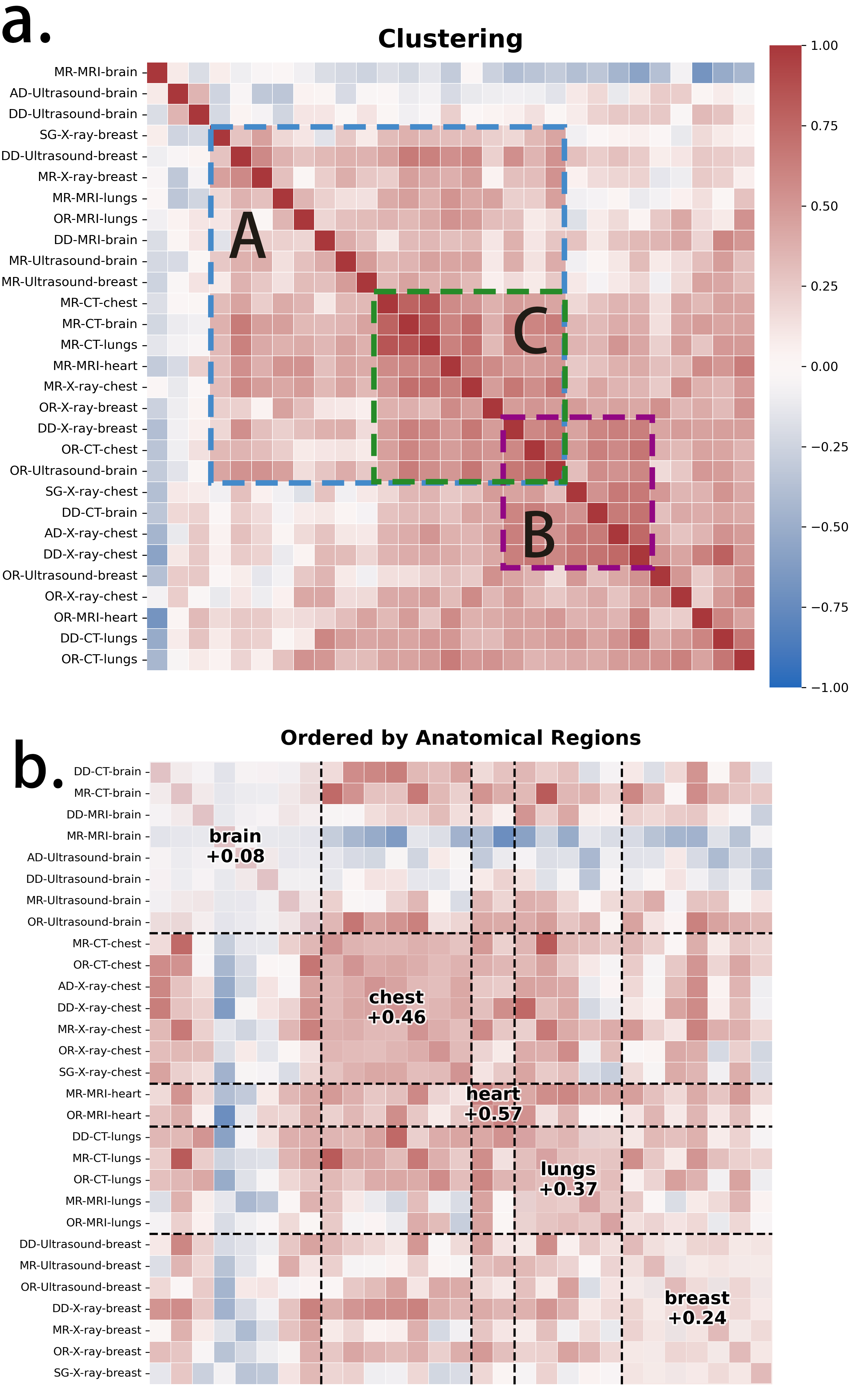}}
    \caption{Correlation Analysis. (a) Hierarchical clustering reveals three functional blocks. (b) Anatomical ordering highlights the brain isolated island effect, where brain tasks show minimal correlation with other regions, reflecting high task heterogeneity.}
    \vspace{-0.5cm}
    \label{fig:correlation}
\end{figure}
\subsection{Settings}

To answer \textbf{RQ1}, we analyze pairwise correlations between competency voxels using Spearman's $\rho$, based on the evaluation results of 33 benchmarked MLLMs. We visualize these relationships using correlation heatmaps and organize them via hierarchical clustering.

\subsection{Potential Skills Factors}
Upon clustering the performance correlation matrix, three distinct functional blocks emerge, as shown in Fig.~\ref{fig:correlation}a. We annotate these groups as Clusters A, B, and C.



\textbf{Capability-Stratified Clusters.}
We observe a clear capability stratification in the correlation structure. \textbf{Cluster A} is dominated by foundational perception tasks (Organ Recognition, Modality Recognition) with high internal coherence ($\bar{\rho}=0.405$), whereas \textbf{Cluster B} groups high-order reasoning tasks (Diagnosis, Grading). Cross-cluster correlations are substantially lower ($\bar{\rho}=0.253$), indicating a persistent gap between perception and reasoning behaviors across models.

\textbf{Resource-Conditioned Sub-Structure.} 
Within this capability scaffold, we observe two salient sub-clustering patterns. First, tasks from specialized niches—such as Breast, as well as specific sub-tasks for Lungs and Brain—form a distinctly separated group (\textbf{Cluster C}). Second, the most coherent portions of both major clusters are dominated by X-ray and CT modalities. This alignment suggests that model consistency is heavily contingent on training signals: stable behaviors develop primarily in \uline{resource-rich} or \uline{readily optimizable} domains.


\subsection{The "Isolated Island" of the Brain}
A striking finding is the profound isolation of brain-related tasks: they show low (and occasionally negative) correlations not only with tasks from other anatomical regions ($\bar{\rho}=0.219$), but even more surprisingly, with one another within the brain domain ($\bar{\rho}=0.089$).
This fragmentation may stem from the intrinsic heterogeneity and complexity of brain-related tasks, or alternatively from brain-focused “score-chasing” models that fail to translate into broader capability gains. To probe these possibilities, we perform additional analyses and find that the effect largely disappears---and sometimes reverses---when brain tasks are stratified by modality, while model-level performance on brain and non-brain tasks remains strongly correlated ($\rho=0.845$). Together, these results favor the former explanation, suggesting that \uline{brain-related tasks impose particularly heterogeneous and modality-dependent capability demands that are difficult to generalize}.

\section{Credibility Analysis}

\begin{figure*}[t]
  \includegraphics[width=\linewidth]{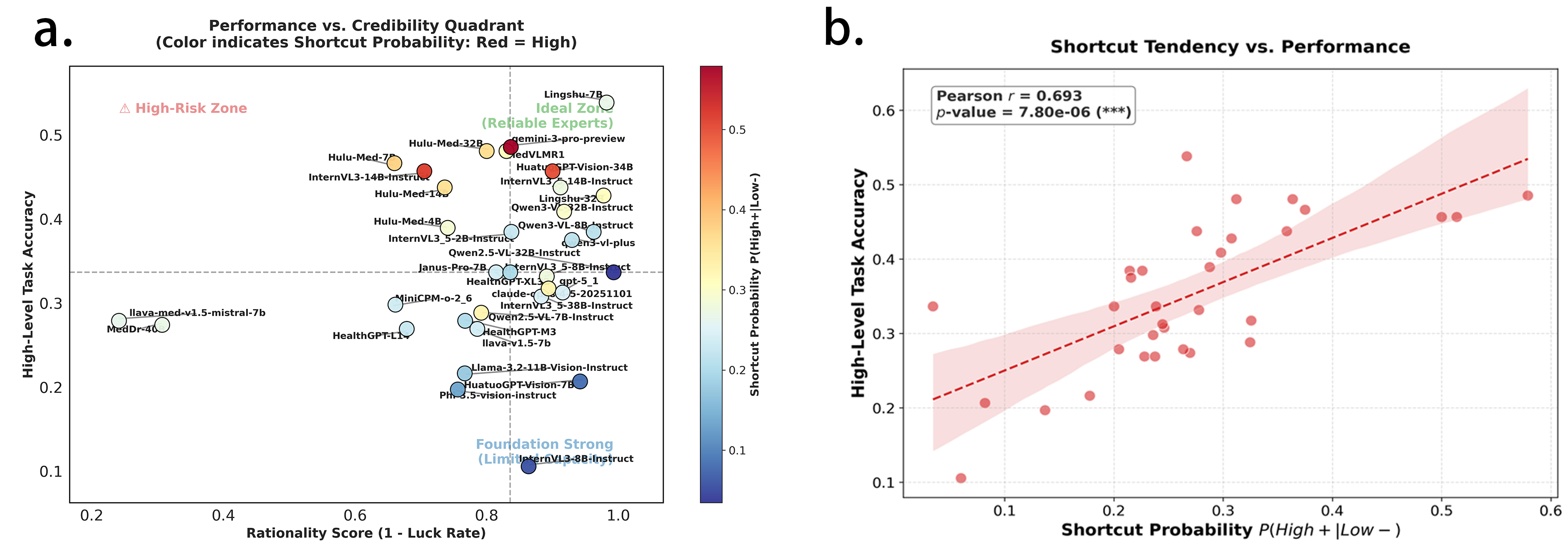}
  \caption {The left panel shows the distribution of models with a quadrant of Rationality score and accuracy. The right panel reveals a strong positive correlation ($r=0.693, p < 10^{-5}$) between Shortcut Probability and accuracy.}
\end{figure*}

\subsection{Settings}
To answer \textbf{RQ2} and operationalize the \textit{Credibility and Consistency} paradigm proposed in \S\ref{subsec:diagnostic_paradigm}, we construct a dedicated subset containing \textbf{300 images}. For each image, we ensure the presence of at least one lower-level or mid-level question (perception or semantic) \emph{and} one higher-level cognition question, so that prerequisite skills and high-level reasoning can be explicitly linked on the same visual evidence. This yields \textbf{900 items} in total, enabling us to directly verify whether a model's diagnostic success is grounded in visual understanding under identical image contexts.

\subsection{Metrics}
We quantify reasoning consistency by paired outcomes $(L,H)$ on the same image, where $L$ denotes the prerequisite task (low/mid-level) and $H$ denotes the cognitive task ($L,H\in\{0,1\}$, with $1$/$0$ indicating correct/incorrect). This yields a $2\times2$ partition:

\textbf{Group A} $(1,1)$ coherent correct; \textbf{Group B} $(1,0)$ broken reasoning; \textbf{Group C} $(0,0)$ both incorrect; \textbf{Group D} $(0,1)$ \textbf{shortcut/incoherent correct}.

Based on the counts $N(\cdot)$, we define two diagnostic metrics:

\textbf{Luck Rate \& Rationality} measures, among instances where the model answers the cognition-level task correctly, the fraction that are not supported by prerequisite correctness. $\mathrm{Rationality Score}$ equal to $1 - \mathrm{LuckRate}$:
\begin{equation}
\label{eq:luck_rate}
\mathrm{LuckRate}=\frac{N(D)}{N(A)+N(D)}.
\end{equation}

\textbf{Shortcut Probability} measures how often a model answers the cognition-level task correctly \emph{given} that it fails the paired lower-level prerequisite, capturing the tendency to ``jump'' to high-level answers without a consistent foundation:
\begin{equation}
\label{eq:shortcut_prob}
\mathrm{ShortcutProb}=\frac{N(D)}{N(C)+N(D)}.
\end{equation}

\subsection{Overall Results}
The Rationality metric reveals a distinct stratification in reasoning reliability (scores ranging 24.1\%--99.3\%). Leading models (e.g., \textit{Lingshu}, \textit{GPT-5.1}, \textit{Qwen}, \textit{HuatuoGPT}) demonstrate superior consistency, achieving Rationality scores above 90\%. In stark contrast, some models specifically fine-tuned for medical domains (e.g., \textit{llava-med-v1.5}, \textit{MedDr-40B}) exhibited alarming behaviors with Luck Rates exceeding 60\%, suggesting their "correct" diagnoses are largely ungrounded guesses or overfits. Notably, \textit{Hulu series} exhibits "hollow" accuracy with suppressed rationality. Crucially, high shortcut dependence persists even in top tiers (e.g., \textit{Gemini-3-pro}, 57.9\%), forming the empirical basis for the \textbf{"Paradox of Specialization"} discussed next.

\subsection{The Paradox of Specialization.}
Intuitively, one would hypothesize an \textit{Ideal Evolution}: as models achieve higher cognitive accuracy, their reliance on shortcuts should diminish, reflecting a transition from guessing to grounded reasoning. 
However, MedRCube exposes a striking contradiction. We observe a strong \textbf{positive correlation} ($r=0.693, p < 10^{-5}$) between high-level accuracy and Shortcut Probability. This supports an alternative \textit{Opportunistic Evolution} hypothesis: \uline{stronger MLLMs are not strictly becoming better observers; rather, they are also becoming better gamblers}.
This implies that performance gains are partly fueled by ungrounded shortcuts, leading to \textbf{evidence-free diagnoses} that are strictly unacceptable in clinical practice. Appendix \ref{sec:appendix_evolution}. provides a clearer elaboration of this discussion.

\section{Related Work}
Text-based benchmarks like MedQA \citep{medqa}, MedMCQA \citep{medmcqa}, and PubMedQA \citep{pubmedqa} evaluate knowledge retention but inherently lack the radiological context critical for real-world diagnosis. To bridge this gap, Radiology VQA emerged, evolving from foundational protocols in VQA-RAD \citep{vqarad}, VQA-Med \citep{vqamed}, and SLAKE \citep{slake} to large-scale automated datasets like PMC-VQA \citep{pmcvqa} and MIMIC-CXR-VQA \citep{mimiccxrvqa}. However, these efforts largely remain single-dimensional, often constrained by noisy alignments or rigid templates that fail to rigorously cover the complexity and seriousness required for high-stakes clinical applications.


Benchmarks have recently transitioned to large-scale, granular evaluation to meet more detailed needs. OmniMedVQA \citep{omnimedvqa} provides a evaluation from modality perspective, while GMAI-MMBench \citep{gmai} leverages lexical trees for fine-grained perception. MedXpertQA \citep{medxpertqa} enhances clinical realism via electronic health records, and ReXVQA \citep{rexvqa} provides a deep dissection of chest X-ray reasoning skills. Collectively, these works expand landscape of assessment breadth and granularity.



These works still overlook the clinical need to assess fine-grained capabilities and measure the credibility of model reasoning.
To address this, MedRCube introduces a multidimensinal architecture to enable in-depth evaluation of MLLMs.

\section{Conclusion}
In this paper, we propose a multidimensional evaluation paradigm to enable fine-grained, in-depth evaluation for medical imaging VQA, and instantiate it with \textbf{MedRCube}. By organizing evaluation in a dense competency space, MedRCube enables a three-level in-depth assessment.
We benchmark \textbf{33} MLLMs on this benchmark, where \textit{Lingshu-32B} achieves the best overall performance, while models from different families and training paradigms exhibit diverse capability profiles across the competency space.
Beyond ranking, our analyses expose multiple weaknesses and blind spots of current models and correlation patterns that suggest potential underlying factors that shape model performance.
We further quantify the credibility of high-level reasoning, revealing large disparities across the models. Notably, we observe a highly significant positive correlation between shortcut behavior and high-level performance, indicating opportunistic evolution and raising concerns about clinically acceptable model development and deployment.


\section*{Limitations}

Our study has following limitations. First, MedRCube is currently confined to the radiological spectrum (e.g., CT, MRI, X-ray), leaving optical imaging domains—such as histopathology and dermatology—unexplored, which restricts the generalization of our findings across the full breadth of medical AI. Second, as a data-driven evaluation, it remains challenging to fully decouple intrinsic model capabilities from performance gains driven by dataset-specific priors or distribution artifacts, potentially introducing confounding factors into our reliability analysis. Third, our credibility metric proxies rationality via hierarchical task consistency rather than explicit visual grounding; we do not currently verify whether high-level diagnoses are supported by pixel-level attention to the correct regions. Finally, despite our systematic construction pipeline and quality assurance, the reliance on automated generation inevitably inherits potential noise or inaccuracies from the original dataset annotations, preventing a complete guarantee of item quality.



\bibliography{custom, models}

\clearpage
\appendix
\section{Large Language Model Usage Declaration}
\label{app:declaration}
We utilized Large Language Models (LLMs) exclusively for code optimization and linguistic refinement. All core scientific contributions—including conceptualization, dataset curation, experimental design, and theoretical framework—were executed solely by the authors.

\section{Data Sources and Statistics}
\label{app:dssss}
\begin{figure*}[t!]
    \centering
    \includegraphics[width=\textwidth]{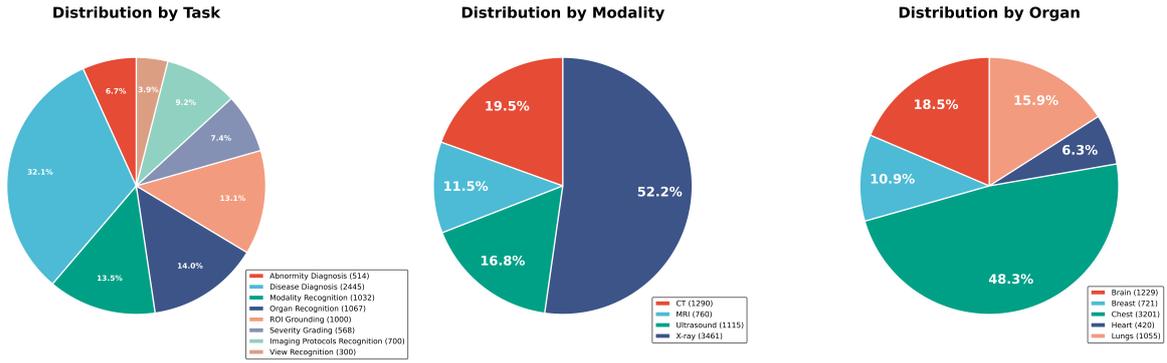}
    \caption{Statistical representation of samples in MedRCube.}
    \label{fig:holisticstatics}
\end{figure*}

To construct a comprehensive holistic benchmark, we curated a total of 35 high-quality datasets spanning four primary imaging modalities. Specifically, the collection consists of 21 X-ray, 6 CT, 6 MRI, and 2 Ultrasound datasets. Recognizing that medical diagnosis is inherently anatomical, we organize these datasets into four distinct anatomical regions: Brain, Heart, Breast, and Chest and Lungs. Table~\ref{tab:medical_datasets} details the sources of these datasets categorized by their respective anatomical targets, while Table~\ref{tab:datasource_stats} reports the per-dataset sample counts. In total, MedRCube comprises 7,626 samples curated from over 30 diverse datasets, providing broad coverage of imaging protocols, clinical contexts, and disease conditions across the above axes. Of these, 1,000 samples are derived from the ROCOv2 dataset \citep{rocov2}. Due to the intrinsic characteristics of this dataset, these samples are incompatible with the voxel system and are thus excluded from it. Consequently, Fig.~\ref{fig:holisticstatics} illustrates the statistical feature distribution of the remaining samples.

\begin{table*}[t!]
\centering
\caption{List of collected medical imaging datasets across different parts used for the benchmark. Total 36 datasets including ROCOv2 used for supplementary analysis, see Appendix~\ref{app:dssss}.}
\label{tab:medical_datasets}
\begin{tabularx}{\textwidth}{lX} 
    \toprule
    \textbf{Part} & \textbf{Collected Datasets} \\
    \midrule
    Brain & Curious 2022 \citep{Curious2022}, CT-ICH \citep{ctich}, BraTS 2020 \citep{brats1, brats2, brats3} \\
    \midrule
    Chest and Lungs & COVIDxCXR-4 \citep{covidxcxr4}, NIH Chest X-ray \citep{nihchest}, SegTHOR \citep{segthor}, COVID-19 Xray Dataset \citep{covidchestxray1, covidchestxray2}, JSRT \citep{jsrt}, Covid-19 Image Dataset \citep{covid19imagedataset}, SIIM-ACR Pneumothorax Segmentation \citep{siimacr}, VinDr-PCXR \citep{vindrpcxr}, COVID-QU-Ex \citep{quex1, quex2, quex3, quex4, quex5}, Pulmonary Chest X-Ray Abnormalities \citep{pulmonartychest1, pulmonartychest2}, COVIDGR \citep{covidgr}, Chest X-Ray Images (Pneumonia) \citep{chestxraypneumonia}, COVID-19 Radiography Database \citep{quex3, quex5}, Chest X-Ray Images with Pneumothorax Masks \citep{siimacr}, CoronaHack \citep{coronahack}, RSNA Pneumonia Detection Challenge \citep{rsna}, COVID19 Pneumonia and Normal Chest X-ray PA Dataset \citep{covid19pa}, Shenzhen Chest X-ray Set \citep{shenzhenset}, Montgomery County CXR Set \citep{shenzhenset}, ChestX-Det \citep{chestxdet}, COVID-CT \citep{covidct}, COVID-19-20 \citep{covid1920}, Learn2Reg2022 \citep{learn2reg}, Finding and Measuring Lungs in CT Data \citep{findingslungsinctdata}, CAD-PE \citep{cadpe} \\
    \midrule
    Breast & BUSI \citep{BUSI}, MIAS Mammography \citep{miasmammo}, CBIS-DDSM \citep{cbisddsm} \\
    \midrule
    Heart & MSD-Heart \citep{msdheart}, M\&Ms \citep{mnms1, mnms2}, HeartSegMRI \citep{heartsegmri}, M\&Ms-2 \citep{mnms2} \\
    \bottomrule
\end{tabularx}
\end{table*}

\begin{table*}[t!]
\centering
\scriptsize
\setlength{\tabcolsep}{3pt}
\renewcommand{\arraystretch}{1.05}
\caption{Detailed sample statistics of data sources in MedRCube (total: 7,626 samples).}
\label{tab:datasource_stats}
\begin{tabularx}{\textwidth}{>{\raggedright\arraybackslash}X r >{\raggedright\arraybackslash}X r}
\toprule
\textbf{Name} & \textbf{\# Samples} & \textbf{Name} & \textbf{\# Samples} \\
\midrule
Curious 2022 & 759 & RSNA Pneumonia Detection Challenge & 79 \\
CT-ICH & 235 & COVID19 Pneumonia and Normal Chest X-ray PA Dataset & 67 \\
BraTS 2020 & 235 & Shenzhen Chest X-ray Set & 79 \\
COVIDxCXR-4 & 79 & Montgomery County CXR Set & 46 \\
NIH Chest X-Ray & 324 & ChestX-Det & 750 \\
SegTHOR & 105 & COVID-CT & 305 \\
COVID-19 Xray Dataset & 43 & COVID-19-20 & 235 \\
JSRT & 56 & Learn2Reg2022 & 105 \\
Covid-19-Image Dataset & 67 & Finding and Measuring Lungs in CT Data & 105 \\
SIIM ACR Pneumothorax Segmentation & 48 & CAD-PE & 305 \\
VinDr-PCXR & 413 & BUSI & 356 \\
COVID-QU-Ex & 56 & MIAS Mammography & 63 \\
Pulmonary Chest X-Ray Abnormalities & 56 & CBIS-DDSM & 302 \\
COVIDGR & 602 & MSD-Heart & 105 \\
Chest X-Ray Images (Pneumonia) & 56 & M\&Ms & 105 \\
COVID-19 Radiography Database & 67 & HeartSegMRI & 105 \\
Chest X-Ray Images with Pneumothorax Masks & 79 & M\&Ms-2 & 105 \\
CoronaHack & 129 & ROCOv2 & 1000 \\
\bottomrule
\end{tabularx}
\end{table*}

\section{Task Definition and Examples}
\label{app:samplesexamples}
\begin{figure*}[t!]
    \centering
    \includegraphics[width=\textwidth]{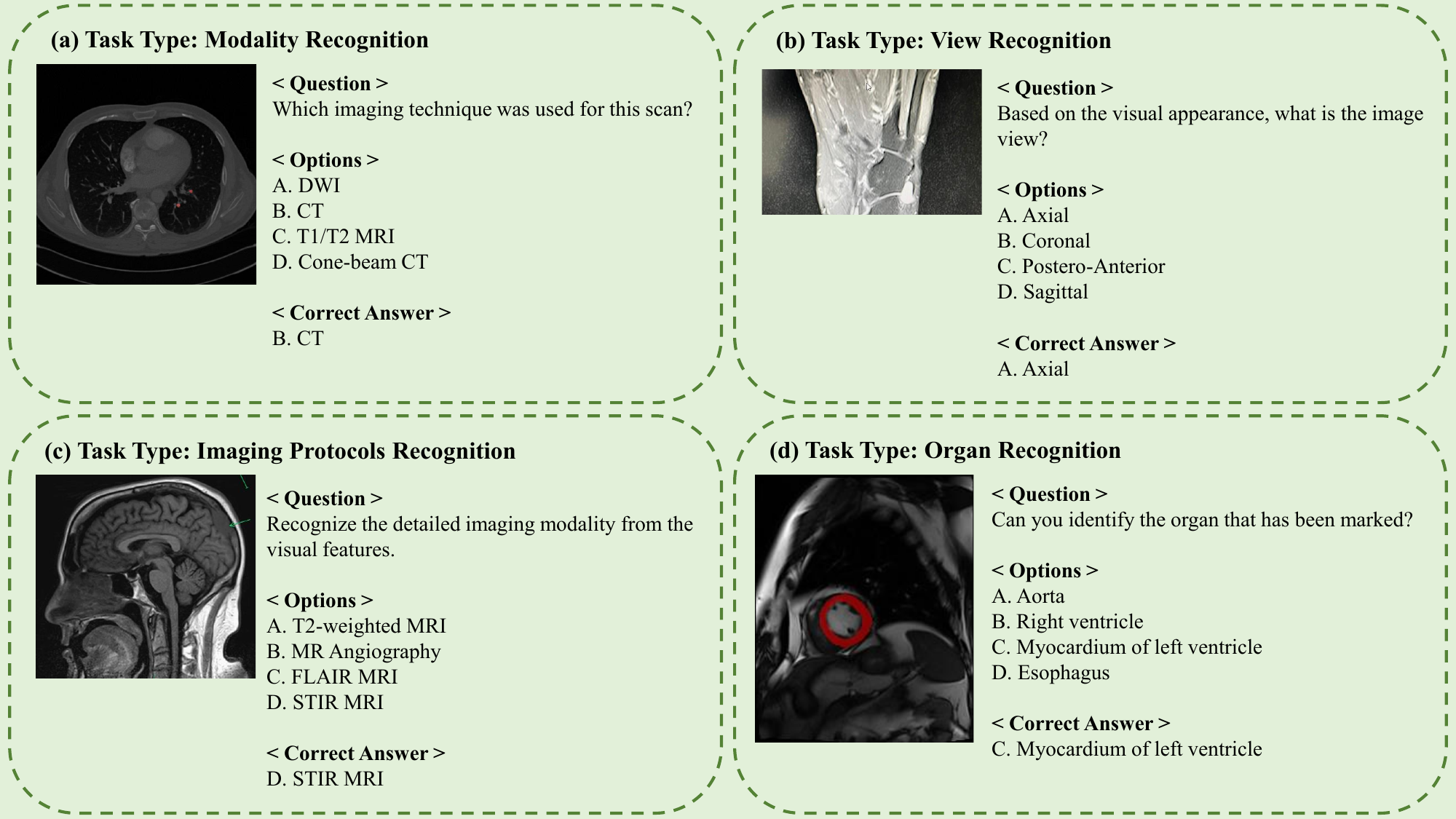} 
    \caption{Illustrative examples of perceptual and semantic tasks in MedRCube. (a) Modality Recognition: Classifies the fundamental imaging modality of a given scan (e.g., CT, MRI) to establish the prerequisite context. (b) View Recognition: Determines the anatomical imaging plane or projection angle (e.g., axial, sagittal). (c) Imaging Protocols Recognition: Identifies specific image acquisition protocols and technical parameters, probing the depth of physical perception. (d) Organ Recognition: Identifies and names salient anatomical structures or specific organs within the image.}
    \label{fig:examples_part1}
    \vspace{0.5cm} 
    \includegraphics[width=\textwidth]{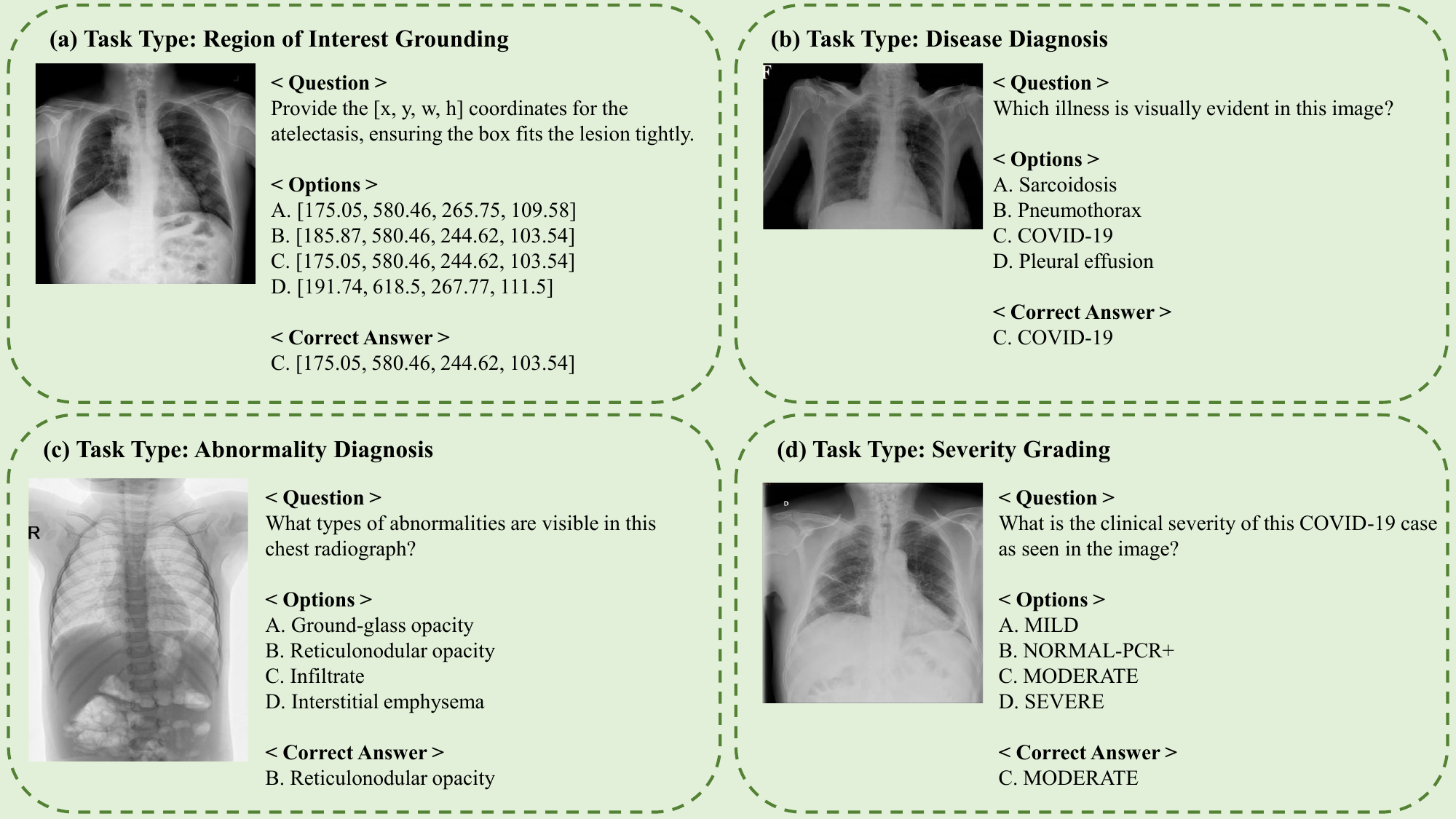} 
    \caption{Illustrative examples of semantic and cognitive reasoning tasks in MedRCube. (a) Region of Interest Grounding: Performs spatial localization of lesions or anomalies to ground semantic concepts onto visual coordinates. (b) Disease Diagnosis: Synthesizes visual evidence to deduce a definitive clinical diagnosis for the observed pathology. (c) Abnormality Diagnosis: Discerns specific radiological signs or phenotypic abnormalities distinct from high-level disease labels. (d) Severity Grading: Assesses the progression stage or severity level of a pathology, requiring precise quantification or ordinal ranking.}
    \label{fig:examples_part2}
\end{figure*}

In this study, we investigate eight distinct tasks, each designed to evaluate a specific dimension of radiological interpretation capabilities:

\begin{itemize}
    \item \textbf{Modality Recognition:} Classifies the fundamental imaging modality of a given scan (e.g., CT, MRI, X-ray), establishing the prerequisite context for subsequent analysis.
    \item \textbf{View Recognition:} Determines the anatomical imaging plane or projection angle (e.g., axial, coronal, sagittal), assessing the model's spatial orientation capabilities.
    \item \textbf{Imaging Protocols Recognition:} Identifies specific image acquisition protocols and technical parameters (e.g., MRI sequences like T1/T2), probing the model's depth of perception regarding medical imaging physics.
    \item \textbf{Organ Recognition:} Identifies and names salient anatomical structures or specific organs within the image, evaluating fine-grained anatomical knowledge.
    \item \textbf{Region of Interest (ROI) Grounding:} Performs spatial localization of pathological lesions or anomalous regions, testing the model's ability to ground semantic concepts onto visual coordinates.
    \item \textbf{Disease Diagnosis:} Synthesizes visual evidence to deduce a definitive clinical diagnosis for the observed pathology.
    \item \textbf{Abnormality Diagnosis:} Discerns specific radiological signs or phenotypic abnormalities (e.g., nodules, effusion) distinct from high-level disease labels, focusing on semiological interpretation.
    \item \textbf{Severity Grading:} Assesses the progression stage or severity level of a pathological condition, requiring precise quantification or ordinal ranking of the disease state.
\end{itemize}
To facilitate a qualitative understanding of the benchmark, we present representative examples for all evaluated tasks. Fig.~\ref{fig:examples_part1} illustrates samples for perceptual and semantic recognition, while Fig.~\ref{fig:examples_part2} displays samples for semantic grounding and cognitive reasoning.

\section{Evaluation Details}
\label{app:evaluationdetails}
Evaluations were conducted via MedEvalKit\footnote{\url{https://github.com/alibaba-damo-academy/MedEvalKit}}. Unless otherwise specified, all experimental settings follow the default configuration provided by MedEvalKit. This section details the metrics and standardized prompts employed to ensure fair evaluation across models.
\subsection{Evaluation Metrics}
We employ Accuracy (ACC) as the primary evaluation metric. The calculation involves two levels of granularity: the overall performance across the entire benchmark and the fine-grained performance across specific dimensions. The Overall Accuracy ($ACC_{overall}$) is calculated using the entire dataset sample set $\mathcal{D}_{all}$:
\begin{equation}
    ACC_{overall} = \frac{N_{correct}(\mathcal{D}_{all})}{N_{total}(\mathcal{D}_{all})} \times 100\%,
\end{equation}
\noindent
where $N_{correct}$ denotes the number of samples correctly answered by the model, and $N_{total}$ is the total number of samples.
To analyze model capabilities in specific domains, we calculate the Type-Specific Accuracy ($ACC_{type}$) for a given subset $\mathcal{D}_{type}$:
\begin{equation}
    ACC_{type} = \frac{N_{correct}(\mathcal{D}_{type})}{N_{total}(\mathcal{D}_{type})} \times 100\%,
\end{equation}
\noindent
where $type$ denotes the specific evaluation category across different axes.

\subsection{Evaluation Models}
A comprehensive roster of all models employed in this assessment is detailed as follows:

\label{app:proprietary}
\textbf{Proprietary Models:} We select the most advanced proprietary models representing the state-of-the-art capabilities, including GPT-5.1~\cite{gpt5_1}, Claude Opus 4.5~\cite{claude_opus_45}, Gemini-3-Pro~\cite{gemini3_pro}, and Qwen3-VL-Plus~\cite{qwen3vl_report}.

\textbf{Medical MLLMs:} We evaluate a comprehensive suite of medical-specific models across different scales. This includes MedVLM-R1~\cite{medvlmr12025}, the HealthGPT series (M3, L14, XL32)~\cite{healthgpt2025}, LLaVA-Med-7B~\cite{li2024llavamed}, HuatuoGPT-Vision (7B, 34B)~\cite{huatuogptvision2024}, Lingshu (7B, 32B)~\cite{lingshu2025}, Hulu-Med (4B, 7B, 14B, 32B)~\cite{hulumed2025}, and MedDr-40B~\cite{meddr2024}.

\textbf{General-purpose MLLMs:} We encompass a wide range of general domain models. This includes the \textbf{InternVL series} (InternVL3 and InternVL3.5, covering 2B, 8B, 14B, and 38B variants)~\cite{internvl35_2025}, the \textbf{Qwen series} (Qwen2.5-VL and Qwen3-VL, covering 7B, 8B, and 32B variants)~\cite{qwen25vl, qwen3vl_report}, Llama-3.2-11B-Vision~\cite{llama3_herd}, Phi-3.5-Vision~\cite{phi35_vision}, Janus-Pro-7B~\cite{januspro2024}, MiniCPM-o 2.6~\cite{minicpmo26}, and LLaVA-v1.5-7B~\cite{liu2023improvedllava}. 

\subsection{Evaluation Implementation}
\label{subsec:evalimpl}

To ensure the exact reproducibility of our benchmark results, we standardize the evaluation pipeline across all model classes. Our implementation is derived from the MedEvalKit framework, and the specific configurations are detailed below.

\textbf{Open-source models.}
This category includes the InternVL3.5 series, Qwen3-VL series, Lingshu series, Hulu-Med series, and other open-weight models evaluated in this work.
\begin{itemize}
    \item \textit{Image preprocessing.} We strictly follow the official inference examples provided in each model's original repository to preprocess input images, ensuring that resolution, normalization, and tokenization match the intended deployment of each model.
    \item \textit{Decoding strategy.} We adopt greedy decoding across all models to minimize sampling variance. The decoding parameters are fixed at temperature $= 0$, $\mathrm{top\_p} = 0.0001$, and $\mathrm{max\_tokens} = 8192$. For models that support an explicit reasoning or thinking mode (e.g., Qwen3-VL), these features are disabled to maintain a standard benchmarking environment.
    \item \textit{Inference framework.} We use vLLM for all supported models to ensure consistent sampling behavior and efficient throughput. For models not yet supported by vLLM (e.g., MedGemma, MedDr), inference is implemented via PyTorch or HuggingFace Transformers, strictly adhering to their official codebases.
\end{itemize}

\textbf{Closed-source models.}
This category includes the proprietary models listed in \S\ref{app:proprietary} (e.g., GPT-5.1, Claude Opus 4.5, Gemini-3-Pro, Qwen3-VL-Plus).
\begin{itemize}
    \item \textit{Execution.} All evaluations are performed via API calls, wrapped with the OpenAI SDK for a unified interface.
    \item \textit{Decoding strategy.} Consistent with the open-source setting, we set temperature $= 0$, $\mathrm{top\_p} = 0.0001$, and $\mathrm{max\_tokens} = 8192$. Any additional reasoning or optimization toggles exposed by the APIs are disabled to ensure comparability.
\end{itemize}

The specific execution configuration for each model is fully contained within our evaluation code and will be released together with the dataset.

\subsection{Text-only Baseline}
\label{app:textonly}
To estimate how much performance can be obtained from textual priors alone, we run an additional \textbf{text-only} setting where models are given the question and options but \emph{no image}. We evaluate a set of representative models spanning proprietary MLLMs, general-purpose open-source LVLMs, and medical LVLMs. Table~\ref{tab:textonly} shows details.

\begin{table*}[t!]
\centering
\caption{Text-only baseline results (no images). Scores are accuracy (\%), and the value in parentheses denotes the absolute drop in Overall accuracy relative to the standard multimodal evaluation.}
\label{tab:textonly}
\resizebox{\textwidth}{!}{%
\begin{tabular}{l|c|cccccccc|cccc|ccccc}
\toprule
\multirow{2}{*}{\textbf{Model}} & \multirow{2}{*}{\textbf{Overall}} & \multicolumn{8}{c|}{\textbf{Task Type}} & \multicolumn{4}{c|}{\textbf{Modality Type}} & \multicolumn{5}{c}{\textbf{Part Type}} \\
\cmidrule(lr){3-10} \cmidrule(lr){11-14} \cmidrule(lr){15-19}
 & & MR & VR & IPR & OR & RG & DD & AD & SG & CT & X-Ray & MRI & US & Brain & Chest & Heart & Lungs & Breast \\
\midrule
GPT-5.1 & 31.52 (21.84$\downarrow$) & 37.98 & 17.33 & 14.00 & 47.20 & 40.10 & 31.34 & 18.12 & 25.43 & 40.70 & 38.41 & 21.58 & 22.33 & 30.57 & 40.14 & 22.14 & 35.55 & 24.13 \\
Gemini-3-Pro & 30.36 (28.99$\downarrow$) & 17.76 & 19.00 & 19.14 & 39.63 & 60.00 & 25.71 & 17.63 & 30.13 & 25.81 & 44.29 & 15.66 & 16.20 & 18.41 & 46.44 & 18.10 & 22.84 & 21.50 \\
InternVL3.5-38B & 34.15 (23.99$\downarrow$) & 24.45 & 27.67 & 24.86 & 41.85 & 51.40 & 35.39 & 10.14 & 32.91 & 38.45 & 41.24 & 27.50 & 19.02 & 18.84 & 42.18 & 31.43 & 40.57 & 28.85 \\
Qwen3-VL-32B & 29.77 (24.66$\downarrow$) & 31.15 & 14.00 & 17.71 & 41.33 & 40.60 & 28.27 & 18.36 & 31.20 & 32.02 & 37.89 & 22.24 & 20.86 & 26.26 & 37.72 & 28.33 & 28.63 & 26.35 \\
HuatuoGPT-Vision-34B & 28.49 (25.53$\downarrow$) & 20.90 & 8.67 & 13.29 & 45.24 & 33.00 & 34.58 & 4.83 & 29.06 & 34.96 & 30.23 & 36.32 & 25.15 & 30.14 & 30.37 & 28.10 & 32.42 & 37.03 \\
Hulu-Med-32B & 27.28 (27.99$\downarrow$) & 28.42 & 28.00 & 27.71 & 25.29 & 25.00 & 28.19 & 27.54 & 27.78 & 24.26 & 29.22 & 25.00 & 26.75 & 24.22 & 28.95 & 26.43 & 24.45 & 29.13 \\
\bottomrule
\end{tabular}%
}
\end{table*}

\section{Construction Details}
\label{app:designdetails}
This section provides further details regarding \S\ref{sec:pipeline}, which were omitted due to space limitations.
\subsection{Image Processing }
\label{subsec:imageprocessing}

Image preprocessing is applied to ensure consistent visualization and input formatting across heterogeneous source datasets.

\textbf{Image protocol.}
For raw medical images that are not provided as pre-rendered slices, image loading and processing are performed using SimpleITK and NiBabel. Modality-specific intensity normalization and color mapping follow the default configurations of the respective libraries. If the original dataset already provides processed slice images, these images are used directly without further modification.

\textbf{Slicing of volumetric images.}
For three-dimensional imaging data, a representative slice is selected for downstream tasks. If pixel-level or region-level annotation masks are available, the slice with the largest annotated area is selected. Otherwise, the central slice of the volume is used. The slice orientation is preserved to remain consistent with the original acquisition direction.

\textbf{Mask visualization.}
Mask annotations are incorporated for visualization when available. Pixel-level segmentation masks are overlaid on the processed image using a semi-transparent red color. For samples annotated with bounding boxes, box coordinates are converted into red rectangular outlines. In cases involving multiple organs or multiple annotation instances, distinct and visually separable colors are used to differentiate different anatomical structures or regions of interest.

\subsection{Question Generation}
\label{subsec:questiongeneration}
To ensure question diversity, we constructed task-specific candidate pools containing a series of question templates. These candidates were rigorously vetted by medical experts to verify their rationality and clinical appropriateness. During dataset construction, the final question for each sample was randomly sampled from this expert-validated pool.

\subsection{Options Design}
\label{subsec:optionsdesign}
We analyze the different question types covered by the dataset and design three systematic and standardized option generation pipelines. Each question is assigned to an appropriate pipeline according to its semantic category and task characteristics, ensuring that all options are accurate, high-quality, and suitable for controlled evaluation.

\paragraph{Disease-oriented Questions}
For questions whose answers correspond to disease entities, candidate distractors are constructed using a two-stage process. Diseases with similar phenotypic characteristics are first identified through similarity matching in the Human Phenotype Ontology (HPO) database. A large language model--based pipeline then generates two groups of candidate diseases or conditions: \textit{similar distractors}, which share overlapping clinical manifestations with the ground-truth answer, and \textit{normal distractors}, which represent clinically plausible but non-matching alternatives to control difficulty. Generated candidates are filtered to remove hierarchical conflicts (e.g., subtypes or parent concepts) and semantic overlap. Final options are obtained through stratified sampling from the pooled candidates.

\paragraph{Abnormality-oriented Questions}
For tasks involving image interpretation or anomaly detection, distractors are derived from abnormal radiological findings observed in the corresponding image. A large language model--based pipeline generates radiological findings with similar imaging characteristics, including \textit{similar distractors} and \textit{normal distractors} corresponding to different difficulty levels. Generated candidates are filtered to avoid hierarchical conflicts (e.g., closely related sub-findings) and semantic redundancy. Final options are selected via stratified sampling to ensure balanced difficulty and diversity.

\paragraph{Predefined Answer Spaces}
For question types with a closed and well-defined answer space, such as binary judgments or limited categorical decisions, options are drawn directly from predefined candidate sets. These candidates are manually verified to ensure completeness, mutual exclusivity, and consistency across samples.

This structured option generation strategy supports diverse question types while maintaining clinical plausibility and controlled difficulty across the benchmark.

\subsection{Standardization and Post-processing}
\label{subsec:standardization}
To ensure terminological consistency and reduce lexical variability in the answer options, a standardized post-processing pipeline is applied. This process focuses on normalizing medical concepts, particularly disease names and radiological terms, to authoritative clinical terminologies.

\textbf{Standardization pipeline.}
All reference terminology entries are encoded and stored in a vector database. For each candidate string to be normalized, a vector similarity search retrieves the top 20 most semantically relevant standardized terms. A large language model then selects the most accurate mapping from these candidates based on semantic and clinical relevance.

\textbf{Terminology sources.}
The pipeline leverages established medical terminologies. RadLex, a radiology lexicon developed by the Radiological Society of North America (RSNA), is used to standardize radiological findings, imaging modalities, and anatomical structures. Disease entities are normalized using ICD-10 and ICD-11 codes, which provide a structured and uniform representation of clinical diagnoses.

This standardization procedure is applied uniformly across all answer options to improve consistency, comparability, and robustness in subsequent benchmarking.

\subsection{Quality Review}
\label{subsec:qualityreview}

To ensure the clinical validity and reliability of the benchmark, all question--answer pairs undergo a structured quality review process adapted from the National Board of Medical Examiners (NBME) item-writing guidelines and tailored to multimodal medical imaging tasks.

\textbf{Question integrity.}
Each question is reviewed independently of its answer options. Clinical accuracy requires that all anatomical, pathological, and radiological terminology be correct, standardized, and error-free. Logical coherence ensures internal consistency without factual or semantic contradictions. Unambiguous phrasing requires concise, grammatically sound, and clearly interpretable wording. Clinical determinism, as a critical criterion, requires that an expert clinician can derive a single definitive answer from the image and question alone, without relying on the answer options.

\textbf{Option set effectiveness.}
Answer options are jointly evaluated with the ground-truth label. Conceptual and grammatical homogeneity requires all options to belong to the same category and share a parallel structure. Mutual exclusivity ensures that options do not overlap semantically. Plausible distractors require incorrect options to be clinically reasonable given the imaging modality and context, while the ground-truth answer remains clearly and uniquely correct.

\subsection{Expert Agreement on LLM-based Quality Review}
\label{subsec:expertagreement}

To verify that the LLM-based quality review yields judgments consistent with human experts, we conducted an agreement study. A Ph.D. candidate in clinical medicine independently annotated 50 randomly sampled question--answer pairs following exactly the same rubric used by the model (see Table~\ref{tab:validation_prompt}, adapted from the NBME item-writing guidelines). We then compared these expert annotations against the judgments produced by GPT-5.1 under the same rubric, and report agreement at two granularities: \textit{sample-level} consistency, which requires the expert and the model to agree on all sub-principles of a sample, and \textit{rule-level} consistency, which measures agreement on each individual principle and sub-principle. As shown in Table~\ref{tab:llm_review_consistency}, rule-level agreement exceeds 90\% on every rule, and sample-level agreement reaches 0.84. These results indicate that the LLM-based quality review is highly aligned with expert judgment, supporting the reliability of using it for large-scale quality control in MedRCube.

\begin{table*}[t!]
    \centering
    \caption{Consistency between GPT-5.1 and a clinical expert on 50 randomly sampled question--answer pairs. Principle~1 covers question integrity (1.1 Clinical \& Anatomical Accuracy, 1.2 Logical Coherence, 1.3 Unambiguous Phrasing, 1.4 VQA Suitability); Principle~2 covers option set effectiveness (2.1 Conceptual \& Grammatical Homogeneity, 2.2 Mutual Exclusivity, 2.3 Plausibility of Distractors). Sample-level consistency requires the all judgments on a sample to match.}
    \label{tab:llm_review_consistency}
    \setlength{\tabcolsep}{6pt}
    \renewcommand{\arraystretch}{1.2}
    \begin{tabular}{l|c|ccccc|cccc}
        \hline
        \textbf{Principle} & \textbf{Sample-level} & \textbf{1} & \textbf{1.1} & \textbf{1.2} & \textbf{1.3} & \textbf{1.4} & \textbf{2} & \textbf{2.1} & \textbf{2.2} & \textbf{2.3} \\
        \hline
        Consistency & 0.84 & 0.94 & 0.98 & 0.98 & 0.94 & 0.98 & 0.90 & 0.94 & 0.96 & 0.96 \\
        \hline
    \end{tabular}
\end{table*}

\section{Explanation of Abbreviations}
Table \ref{tab:abbre} outlines the abbreviations and corresponding full designations for the specific medical tasks integrated into MedRCube.
\label{app:abbre}
\begin{table}[H] 
    \centering
    \caption{Abbreviations and Full Names of Tasks}
    \label{tab:abbre}
    \begin{tabularx}{\linewidth}{X|c}
        \hline
        Abbreviation & Full Name \\
        \hline
        DD & Disease Diagnosis \\
        MR & Modality Recognition \\
        AD & Abnormality Diagnosis \\
        SG & Severity Grading \\
        OR & Organ Recognition \\
        RG & Region of Interest Grounding \\
        IPR & Imaging Protocols Recognition \\
        VR & View Recognition \\
        \hline
    \end{tabularx}
\end{table}

\section{Prompts Template}
\subsection{Prompt Templates for Options Generation}
We detail the prompt engineering strategy for distractor generation. Strictly adhering to National Board of Medical Examiners (NBME) guidelines, we categorize generation tasks into Disease-Oriented and Abnormality-Oriented types. For both categories, our templates are designed to mitigate elimination shortcuts by constructing two distinct classes of distractors: similar distractors and normal distractors. Table~\ref{tab:prompt_disease} and Table~\ref{tab:prompt_abnormality} report on the prompt design and purpose for two tasks.

\begin{table*}[t!]
    \centering
    \renewcommand{\arraystretch}{1.3}
    \caption{Prompt templates for disease-oriented task. Braced placeholders are dynamic variables instantiated based on sample metadata: \texttt{\{disease\}} denote the ground-truth diagnosis; and \texttt{\{num\_distractors\}} controls the required number of options.}
    \label{tab:prompt_disease}
    
    \begin{tabular}{|p{0.2\linewidth}|p{0.75\linewidth}|}
        \hline
        \multicolumn{2}{|c|}{\cellcolor{gray!15}\textbf{Prompt Templates for Disease-Oriented Task}} \\
        \hline
        \textbf{Type} & \textbf{Prompt Content} \\
        \hline
        \textbf{Similar Distractors} & \textit{Prompt:} Generate \{num\_distractors\} diseases similar to `\{disease\}' but distinct and incorrect. Focus on diseases with similar symptoms or pathophysiology. \newline
        \textit{Purpose:} Leverage the Human Phenotype Ontology (HPO) to generate difficult, clinically confusing options requiring fine-grained differentiation. \\
        \hline
        \textbf{Normal Distractors} & \textit{Prompt:} Generate \{num\_distractors\} diseases from the same anatomical region as '\{disease\}' with similar disease category. \newline
        \textit{Purpose:} Serve as a control group using cross-region diseases to regulate difficulty and validate basic anatomical alignment. \\
        \hline
    \end{tabular}

    \vspace{2em} 

    \caption{Prompt templates for abnormality-oriented task. Braced placeholders represent dynamic variables instantiated from sample metadata: \texttt{\{finding\}} denote the specific radiological finding; \texttt{\{region\}} specify the anatomical location; \texttt{\{modality\}} provides the necessary imaging modality context; and \texttt{\{num\_distractors\}} governs the required quantity of generated options.}
    \label{tab:prompt_abnormality}

    \begin{tabular}{|p{0.2\linewidth}|p{0.75\linewidth}|}
        \hline
        \multicolumn{2}{|c|}{\cellcolor{gray!15}\textbf{Prompt Templates for Abnormality-Oriented Task}} \\
        \hline
        \textbf{Type} & \textbf{Prompt Content} \\
        \hline
        \textbf{Similar Distractors} & \textit{Prompt:} Generate \{num\_distractors\} imaging findings that could be confused with `\{finding\}' in \{modality\} of \{region\}. \newline
        \textit{Purpose:} Generate region-specific, plausible abnormalities to provide high-difficulty confounders for the target finding. \\
        \hline
        \textbf{Normal Distractors} & \textit{Prompt:} Generate \{num\_distractors\} common imaging findings from \{region\} that could appear in differential diagnosis. \newline
        \textit{Purpose:} Serve as a control group using cross-region or cross-modality findings to regulate difficulty and test context awareness. \\
        \hline
    \end{tabular}
    \vspace{2em}
    \caption{The unified prompt template used for model evaluation. The placeholder \texttt{\{question\}} represents the clinical query of the test sample, and \texttt{\{options\}} denotes the candidate choices provided for that sample.}
    \label{tab:eval_prompt}
    \begin{tabular}{|p{0.96\linewidth}|}
        \hline
        \cellcolor{gray!15}\textbf{Prompt Templates for Model Evaluation} \\
        \hline
        Question: \{question\} \newline
        Options: \{options\} \newline
        Answer with the option's letter from the given choices directly. \\
        \hline
    \end{tabular}
\end{table*}

\subsection{Evaluation Prompt}
To ensure a standardized assessment, we utilize a unified zero-shot prompt template for all participating models. As illustrated in Table~\ref{tab:eval_prompt}, the prompt is strictly structured to guide the model to answer medical question.

\subsection{Quality Review Prompt}
The prompts used for question-and-answer pair checking are shown in Table~\ref{tab:validation_prompt}.

\section{Evolutionary Hypotheses and Diagnostic Predictions}
\label{sec:appendix_evolution}

\subsection{Preliminaries and Notation}
\label{subsec:appendix_prelim}

High-level (cognition-level) accuracy and Shortcut Probability are defined as:
\begin{equation}
Acc_H = \frac{N(A)+N(D)}{N(A)+N(B)+N(C)+N(D)},
\end{equation}
\begin{equation}
P_{SC} = \frac{N(D)}{N(C)+N(D)}.
\end{equation}

Importantly, $Acc_H$ conflates two qualitatively distinct sources of correct cognition-level answers: grounded correctness ($N(A)$) and shortcut-based correctness ($N(D)$). In contrast, $P_{SC}$ isolates ungrounded success by conditioning on prerequisite failure ($L=0$).

\subsection{Hypothesis}

\paragraph{H0: Ideal Evolution.}
As models scale, improvements in cognition-level accuracy are primarily driven by stronger perceptual and semantic grounding, yielding more causally valid reasoning chains. Reliance on shortcut-based guessing diminishes as ambiguity is resolved.

\textit{Expected dynamics.}
\begin{equation}
\Delta N(A) > 0,\qquad \Delta N(D) \le 0.
\end{equation}

\noindent \textbf{Expected Prediction:}
Since $P_{SC}$ is monotonically increasing in $N(D)$, and H0 requires $\Delta N(D)\le 0$, improvements in $Acc_H$ should not be accompanied by higher shortcut reliance. Thus, H0 predicts a non-positive or weak association:
\begin{equation}
\mathrm{Corr}(Acc_H, P_{SC}) \le 0.
\end{equation}

\paragraph{H1: Opportunistic Evolution (Pattern Exploitation).}
As models scale, cognition-level accuracy improves through both enhanced grounding and increasingly effective exploitation of statistical shortcuts that bypass prerequisite understanding.

\textit{Expected dynamics.}
\begin{equation}
\Delta N(A) > 0,\qquad \Delta N(D) > 0.
\end{equation}

\noindent \textbf{Expected Prediction:}
Under H1, growth in shortcut-based correctness ($\Delta N(D)>0$) directly increases $P_{SC}$. As a result, models with higher cognition-level accuracy are expected to exhibit higher Shortcut Probability, yielding:
\begin{equation}
\mathrm{Corr}(Acc_H, P_{SC}) > 0.
\end{equation}

\begin{table*}[t!]
    \caption{Prompt of Quality Review. This prompt is adapted from the National Board of Medical Examinations (NBME) guidelines for item writing and has been tailored for multimodal medical imaging tasks to provide rigorous review of question items and options.}
    \label{tab:validation_prompt}
    \begin{tabular}{|p{0.96\linewidth}|}
        \hline
        \cellcolor{gray!15}\textbf{Prompt Templates for Quality Review} \\
        \hline
        SYSTEM\_PROMPT = """ \newline
        You are a Senior Medical Content Analyst. \newline
        You are performing a **TEXT-ONLY pre-validation** of a medical VQA dataset. \newline
        You will **NOT** be provided with any images. Your evaluation must be based solely on the textual content (question, options, etc.) provided. \newline
        Your objective is to ensure the clinical integrity and educational value of the VQA samples by performing a critical review of their textual components. \newline
        You will evaluate each sample based on two stringent principles. You MUST evaluate and provide a 0 (Fail) or 1 (Pass) for EACH individual sub-principle (e.g., 1.1, 1.2, ... 2.3). \newline
        \newline
        **Principle 1: Question Integrity (Text-Only Focus)** \newline
        * **1.1. Clinical \& Anatomical Accuracy:** All terminology (e.g., anatomical structures, pathologies, findings) in the question must be accurate, standard, and free of error. \newline
        * **1.2. Logical Coherence:** The question must be internally consistent and free of factual or logical self-contradictions. \newline
        * **1.3. Unambiguous Phrasing:** The question must be concise, grammatically sound, and semantically unambiguous. \newline
        * **1.4. CRITICAL - VQA Suitability (Image-Dependence \& Objectivity):** \newline
        The question's phrasing must be structured as a valid VQA query. This means it must satisfy TWO conditions: \newline
        * **a) Image-Dependent:** The question MUST require visual information (from the hypothetical image) to be answered. \newline
        (FAIL if the question is general knowledge, e.g., "What is pneumonia?" or "Define lung."). \newline
        * **b) Objective \& Factual:** The question must seek a factual, objective answer based on visual content. \newline
        (FAIL if the question is subjective, e.g., "Is this a high-quality image?", "What do you think of this finding?"). \newline
        * **ALLOWED:** Questions that directly reference the (hypothetical) image or its parts are valid, even if they seem open-ended. \newline
        (PASS example: "Please observe this ultrasound image, what problem is shown in the area marked by the red box?") \newline
        (PASS example: "What abnormality is present in the lower right quadrant?") \newline
        \newline
        **Principle 2: Option Set Effectiveness (Text-Only Focus)** \newline
        * **2.1. Conceptual \& Grammatical Homogeneity:** All options (A, B, C, D) must belong to the same category (e.g., all diagnoses, all anatomical parts, all 'yes/no' judgments) and share a parallel grammatical structure. \newline
        * **2.2. Mutual Exclusivity:** Options must be distinct and not overlap (e.g., "Pneumonia" and "Bacterial Pneumonia" as separate options constitutes a failure). \newline
        * **2.3. Plausibility of Distractors:** All incorrect options (distractors) must be plausible, reasonable "foils" given the clinical context (modality, anatomical regions). The `gt\_answer` must be unequivocally the *best* and most correct answer among the choices. \newline
        You MUST return your evaluation as a single, minified, valid JSON object. \newline
        NO explanatory text, markdown (``json ... ''), or any other characters outside the JSON structure are permitted. \newline
        """ \\
        \hline
    \end{tabular}
\end{table*}

\end{document}